\documentclass[a4paper,fleqn]{cas-dc}

\usepackage{amsmath,amssymb}
\usepackage{booktabs}
\usepackage{subcaption}
\usepackage{url}

\usepackage[numbers]{natbib}
\usepackage{pdfpages}

\ExplSyntaxOn
\cs_set:Npn \__first_footerline:
{
  \group_begin:
  \small
  \sffamily
  \__short_authors: :~
  { \rmfamily \itshape Preprint~ (under~ review) }
  \group_end:
}
\ExplSyntaxOff

\newcommand{\CM}{C\,+\,M}
\newcommand{\CnM}{C\,+\,\={M}}
\newcommand{\nCM}{\={C}\,+\,M}
\newcommand{\nCnM}{\={C}\,+\,\={M}}

\begin{document}

\shorttitle{Vision--language classification of TdP risk in ECG images}

\shortauthors{D. Labrador Alonso and A. Alahmadi}

\title[mode=title]{Vision--language classification of Torsades de Pointes risk from pseudo-coloured ECG images: do T-wave morphology and colour context help or hinder?}

\author[1]{Dario Labrador Alonso}
\credit{Methodology, Software, Formal analysis,
  Investigation, Data curation, Writing -- original draft,
  Writing -- review \& editing}
\affiliation[1]{organization={School of Computing, Newcastle University},
            city={Newcastle upon Tyne},
            postcode={NE4 5TG},
            country={United Kingdom}}

\author[1]{Alaa Alahmadi}
\cormark[1]
\ead{alaa.alahmadi@newcastle.ac.uk}
\credit{Conceptualization, Resources, Data curation, Supervision,
  Writing -- review \& editing}

\cortext[1]{Corresponding author.}

\begin{abstract}
Contextual prompting is widely presented as a cheap route to better performance
from vision models, but the evidence comes largely from benchmarks in which an
attribute named in a prompt is also present in the image, confounding the two
channels. We investigate the significance of context for a semantically complex
biomedical signal classification task using a domain-adapted vision--language
model, BiomedCLIP. Building on our previous work that renders electrocardiogram
(ECG) signals as colour-enhanced images to flag Torsades de Pointes risk using
the QT-nomogram and pseudo-colouring technique, we cross four context strategies
with four T-wave morphology label sets, evaluated zero-shot and with a
class-balanced linear probe, under five-fold stratified cross-validation on
9{,}789 images. Zero-shot inference is unusable. Once trained, the
colour-enhanced images outperform the colourless ones on every metric: balanced
accuracy 0.903 versus 0.825, sensitivity 0.914 versus 0.827, specificity 0.892
versus 0.823, ROC-AUC 0.955 versus 0.888 and Matthews correlation coefficient
(MCC) 0.437 versus 0.311. Crossing colour-in-image with colour-in-prompt
separates the confounded channels: the image contributes $+0.126$ MCC
($+0.078$ balanced accuracy, $+0.067$ ROC-AUC) whether or not the prompt names
colour, whereas naming it changes nothing under any metric. Context is
two-sided: the best prompt set gains 0.008 MCC where colour is present and 0.016
where it is not, while the worst loses 0.017 and 0.013. For a signal rendered as
an image, the leverage lies in how the signal is drawn rather than in how it is
described --- and where the endpoint is an interval measurement, prompts naming
T-wave morphology describe the wrong quantity.

\end{abstract}


\maketitle

\section{Introduction}

\noindent Contextual prompting --- supplying a vision--language model with text describing attributes of the scene --- is a central strand of computer-vision research
\citep{An2024,Godec2010,Wang2023}, and a contested one. One body of work treats
contextual attributes as a meaningful contributor to recognition
\citep{An2024,Godec2010,Ge2023,Wang2023}; another argues that context is largely
redundant when appearance information is strong and valuable only when the
visual signal is weak \citep{Wolf2006}. An intermediate position holds that the
form context takes matters as much as its presence \citep{Lu2024,ZhangICL2023}.
Almost all of this evidence comes from benchmarks of common objects
\citep{Lin2014,Deng2009}, where the target is conspicuous.

Two things make that evidence hard to interpret. First, on such benchmarks the
attribute named in the prompt is also visible in the image, so when performance
improves it is not possible to say which channel did the work. Second, the
labels are object identities, which prompts can name directly; many applied
tasks are not like this.

Torsades de Pointes (TdP) risk assessment is not like this. TdP
\citep{Dowd2014} is a potentially fatal polymorphic ventricular tachycardia
associated with drug-induced prolongation of the QT interval. Whether a QT value
is dangerous depends on heart rate, so the decision is made with a QT nomogram,
which plots QT against heart rate and separates an at-risk region above a
threshold line \citep{Chan2007}. The label is thus a statement about interval
duration relative to a rate-dependent threshold --- not about waveform shape.
Alahmadi et al.\ \citep{Alahmadi2021,Alahmadi2023} built an explainable algorithm
around this rule and made its decision visible by pseudo-colouring: the area
under the repolarisation curve is filled with a colour scale anchored so that
the nomogram's threshold for the beat's own heart rate is rendered dark orange,
with anything beyond it red \citep{Alahmadi2020}. The same beats are also
available without colour.

This dataset supports a controlled test that the usual benchmarks do not. Colour
can be present or absent in the \emph{image} (the two renderings) and named or
not named in the \emph{prompt} (a colour label set), and crossing the two
separates the confounded channels: if the discriminative work is done by the
text, performance should track whether colour is named; if by the image, it
should track whether colour is present. Separately, because the reference
algorithm was deliberately built to be robust to T-wave morphology --- QT-prolonging
drugs distort T-wave shape, which is what invalidates automated QT measurement
\citep{Alahmadi2021,Vicente2015} --- morphological description is a \emph{proxy}
for the label rather than a statement of it.

We therefore ask: (RQ1) does morphological context help or hinder a classifier
whose labels are morphology-independent, and does that depend on whether the
label-defining cue is visible? (RQ2) when a vision--language model appears to use
a visual attribute named in a prompt (e.g., colour), is that attribute read from the image or carried by the text?

\paragraph{Contributions:}
We (1) formulate nomogram-defined TdP risk detection from pseudo-coloured ECG
images as a context-driven vision--language problem using BiomedCLIP
\citep{BiomedCLIP}; (2) evaluate a factorial design of four context strategies
$\times$ four morphology label sets, zero-shot and trained, under five-fold
stratified cross-validation on both renderings; (3) show that the colour signal
is read from the image and that naming it in text adds nothing, in either
rendering and under every metric; and (4) show that morphological context helps
only where the label-defining overlay is absent, that single-word morphological
categories are the weakest prompts tested, and that the gains are
operating-point gains rather than improvements in separability.

\section{Related work}

\noindent Wang and Zhu \citep{Wang2023} distinguish \emph{appearance} context (colours and shapes) from \emph{semantic} context (meaning carried by a description), and
separate levels of use from prior knowledge through global to local; Torralba
et al.\ \citep{Torralba2003} make a parallel case for spatial context. Their
taxonomy maps onto our design: the colour overlay and T-wave shape are
appearance context, the prompts are semantic context. Wang and Zhu also warn
that bad context can hurt as much as good context helps. Against this, Wolf and
Bileschi \citep{Wolf2006} argue that high-level context is redundant when
appearance evidence is strong and useful only when it is weak, and Zhang et al.\
\citep{ZhangICL2023} show that different in-context examples produce markedly
different results.

PerceptionCLIP \citep{An2024} is the closest methodological antecedent: it
enriches CLIP \citep{Radford2021} prompts with contextual attributes and uses
similarity scores to approximate the conditional probabilities needed for
perception-based classification, reporting gains across datasets from ImageNet
\citep{Deng2009} to CelebA \citep{Liu2015} and CUB-200 \citep{Wah2011}, while
noting CLIP's sensitivity to the exact text used --- the motivation for our four
label sets. Alternatives externalise reasoning to a language model
\citep{Berrios2023,Sakai2026}. Ge et al.\ \citep{Ge2023} and Lu et al.\
\citep{Lu2024} find that prompt semantics and prompt \emph{structure} govern how
strongly text binds to visual entities, and Zhang et al.\
\citep{ZhangGroundHOG2024} use region-referring prompts with multimodal language
models \citep{Yin2024}. Godec et al.\ \citep{Godec2010} introduce virtual classes
that collapse to one decision, which we adopt for our multi-prompt label sets.

In medical imaging, Zhao et al.\ \citep{Zhao2025} find CLIP-driven applications
underperform because CLIP is pre-trained on natural images, the usual remedy
being domain pre-training. We therefore use BiomedCLIP \citep{BiomedCLIP},
pre-trained on fifteen million biomedical image--text pairs; Cardiac-CLIP
\citep{Hu2025} is trained on 3D cardiac CT and is too distant from a rendered
ECG. On the clinical side, the QT nomogram \citep{Chan2007} defines our labels,
pseudo-colouring \citep{Alahmadi2020,Alahmadi2021,Alahmadi2023} produces the
coloured rendering, and Lester and Paglialunga \citep{Lester2025} supply the
morphological vocabulary used in our prompts.

What is missing from this literature is a setting in which the strength of the
appearance cue can be varied deliberately while labels and signals are held
fixed. That is what follows.

\section{Materials and methods}\label{sec:methods}

\subsection{Data and labels}

\noindent The collection comprises 9{,}792 PNG images, each showing a single beat spanning one R-to-R interval, in two subsets: \emph{pseudo-coloured} and
\emph{colourless}. The underlying signals derive from a randomised clinical
trial of four QT-active drugs and placebo in healthy subjects
\citep{Johannesen2014,pollard_physionet_2026}, the corpus used to develop the source algorithm of pesduoc-poluring the ECGS  \citep{Alahmadi2021}.

The classification of TdP risk is based on the QT-nomogram, using the clinical trials reported values of the uncorrected QT interval (QT) that is read against the observed heart rate on the nomogram of Chan et al.\ \citep{Chan2007}, and the beat is labelled \emph{high risk} if the pair falls on
or above the threshold line and \emph{low risk} otherwise. High risk is the
positive class. Analyses use the images the pipeline processed successfully:
5{,}035 pseudo-coloured (181 high-risk, 4{,}854 low-risk) and 4{,}754 colourless
(181 high-risk, 4{,}573 low-risk), 9{,}789 in total. Both subsets are strongly
imbalanced, at 3.6--3.8\% positives. Images were opened with Pillow
\citep{Pillow}, converted to RGB and passed through the BiomedCLIP validation
transform ($224\times224$); no other preprocessing was applied, so any
difference between conditions is attributable to the prompts.

\subsection{The pseudo-colour overlay}\label{sec:pseudocolour}

\begin{figure*}[htbp]
\centering
\includegraphics[width=0.6\textwidth]{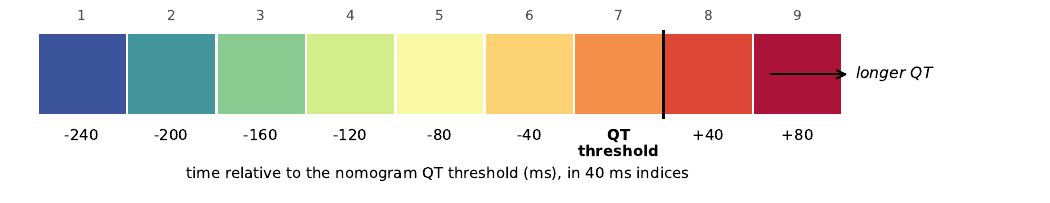}
\caption{The nine-index pseudo-colour scale, with swatches sampled from the
dataset images. Each index spans 40\,ms. Index seven is the QT value the
nomogram gives as the risk threshold for the beat's own heart rate; indices
eight and nine lie beyond it. Because the anchor moves with heart rate the scale
encodes a rate-normalised judgement, and because the window is fixed the whole
scale appears on every beat \citep{Alahmadi2020,Alahmadi2021}.}
\label{fig:scale}
\end{figure*}

\noindent Sequential time is added to each beat as a third dimension from approximately
the onset of the Q wave, and the area under the curve is filled with a
spectrum-approximation colour sequence in which each successive 40\,ms --- one
small square on the standard ECG grid --- is mapped to one of nine colour
indices \citep{Alahmadi2020,Alahmadi2021}. The scale is anchored to the nomogram,
not to the beat: the QT value the nomogram gives as the risk threshold
\emph{for that beat's heart rate} is index seven, rendered dark orange; indices
eight and nine (40 and 80\,ms beyond) are red and dark red; indices six to one
run back through orange, yellow, lime, green, blue and purple
(Fig.~\ref{fig:scale}).

\begin{figure*}[htbp]
\centering
\begin{subfigure}{\textwidth}
  \centering
  \includegraphics[width=0.30\linewidth]{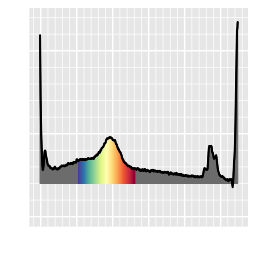}\hspace{0.04\linewidth}
  \includegraphics[width=0.30\linewidth]{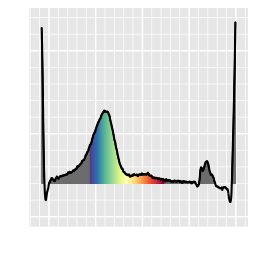}
  \caption{High-risk (left) and low-risk (right) beats \emph{with} the pseudo-colour overlay.}
\end{subfigure}

\vspace{0.8em}

\begin{subfigure}{\textwidth}
  \centering
  \includegraphics[width=0.30\linewidth]{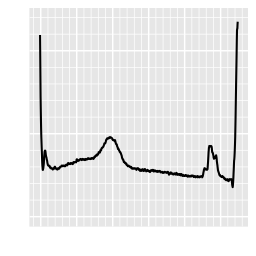}\hspace{0.04\linewidth}
  \includegraphics[width=0.30\linewidth]{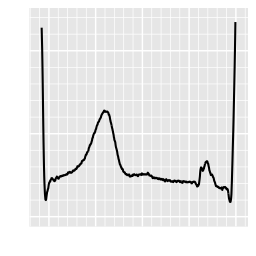}
  \caption{The same ECGs \emph{without} pseudo-colour.}
\end{subfigure}
\caption{Example of single heartbeat ECG. In the high-risk example the repolarisation is still
elevated at the dark-orange and red indices; in the low-risk example the T wave
has returned to baseline while still in the cool part of the scale.}
\label{fig:ecgs}
\end{figure*}

Three properties matter. (i) The coloured window is fixed at
$9\times40=360$\,ms regardless of the beat's QT, so every image of either class
contains the whole scale; the classes are distinguished by \emph{where the
T wave sits on it} (Fig.~\ref{fig:ecgs}). (ii) Because the anchor moves with
heart rate, the overlay presents an already rate-normalised judgement, whereas
recovering the same decision from the colourless rendering requires recovering
both interval and rate. (iii) The overlay is a rendering of the labelling rule,
so the coloured condition asks whether the model can read a clinical
explainability overlay while the colourless condition asks whether it can infer
the decision from waveform appearance. The performance gap between subsets
should be read in that light rather than as a difference in task difficulty.

\subsection{Prompts and context strategies}\label{sec:labels}

\begin{figure*}[htbp]
\centering
\includegraphics[width=\textwidth]{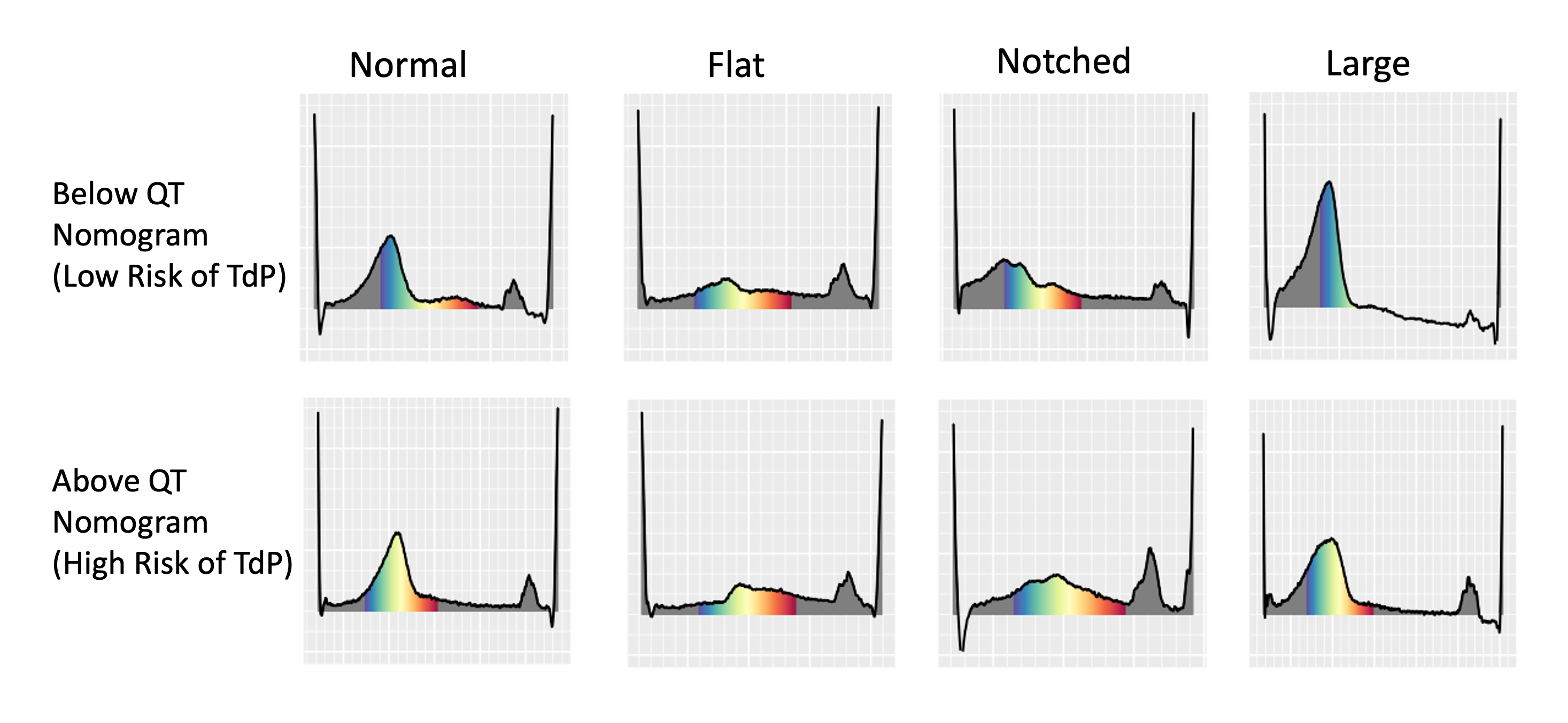}
\caption{Examples of pseudo-coloured ECGs with the T-wave morphologies named in label set L2 and described in L3. ``Large'' denotes a T-wave taller or broader than usual.}
\label{fig:morph}
\end{figure*}

\noindent Four morphology label sets of increasing structural detail were
defined as text prompts. \textbf{No Label} (2 prompts): ``Normal'';
``Abnormal''. \textbf{L1} (2): ``Normal T wave''; ``Abnormal T wave''.
\textbf{L2} (6): ``Normal'' plus five single-word abnormality categories ---
``Flat'', ``Large'', ``Inverted'', ``Biphasic'' and ``Notched'' --- supplied
exactly as listed, with no accompanying description (Fig.~\ref{fig:morph}).
\textbf{L3} (10): descriptive sentences naming both the abnormality and its
reason, e.g.\ ``Abnormal large T-wave, broader and larger than initial
signal''; ``Abnormal T-wave with a deep inversion''. Following Godec et al.\
\citep{Godec2010}, every non-\emph{Normal} prompt resolves to the positive
class, keeping the decision binary while directing attention to different
properties of the trace. Morphology is a proxy channel: QT-prolonging drugs
alter T-wave shape in drug-specific ways \citep{Vicente2015,Lester2025}, so
morphological description carries information about drug exposure and hence
indirectly about prolongation, but never defines the label.

A separate two-prompt \textbf{colour} set names the overlay: ``a T wave with a
high amount of blue/green'' (low risk) and ``a T wave with a high amount of
orange/red'' (high risk). Because the colouring fills the area under the curve,
a prolonged T wave is literally filled with warm colour, so localising the
description to the T wave makes the prompt approximately correct; what it does
not express is the threshold structure of the scale.

Crossing the two sets gives four \textbf{context strategies}: \nCM{} (morphology
only), \CnM{} (colour only), \CM{} (both) and \nCnM{} (neither, the no-context
baseline). Colour strategies were retained on the colourless subset as a
negative control. Crossing strategies with renderings gives the $2\times2$
design --- colour present or absent in the image, named or not named in the
prompt --- that answers RQ2.

\subsection{Model, fusion and training}

\noindent BiomedCLIP \citep{BiomedCLIP} was loaded from the Hugging Face Hub
\citep{HuggingFace} through \texttt{open\_clip} \citep{OpenCLIP} and used frozen
throughout; no weights are updated. Image and text embeddings are
$L_2$-normalised, and scaled cosine similarities are softmaxed into two
distributions per image: $P_m$ over morphology prompts and $P_c$ over the two
colour prompts. The \CM{} strategy fuses them with a weight
$W = \alpha P + (1-\alpha)$ computed from the colour distribution
($\alpha=0.5$, fixed a priori): prompts containing ``Normal'' are multiplied by
$W_{\mathrm{blue/green}}$ and ``Abnormal'' prompts by $W_{\mathrm{orange/red}}$,
and the result is renormalised. With $\alpha=0.5$ the factor lies in
$[0.5,1.0]$, so colour can modulate but never override morphology.

Experiment~1 uses these distributions directly for zero-shot prediction.
Experiment~2 concatenates the normalised image embedding with the strategy's
probability vector \citep{Cheng2016},
$X_{\mathrm{concat}} = [\,X_{\mathrm{emb}},\,P_{\mathrm{strategy}}\,]$, and
fits a logistic-regression probe \citep{Pedregosa2011} with
\texttt{class\_weight="balanced"}, following the linear-probe protocol of
Radford et al.\ \citep{Radford2021}. Under \nCnM{} the appended block carries no
contextual information, making the strategy comparison a clean test of what
prompts contribute. Five-fold stratified cross-validation preserves class
proportions ($\approx36$ positives per fold).

\subsection{Evaluation}

\noindent Metrics are derived from the confusion matrix \citep{IBMConfusion}: sensitivity and specificity \citep{McKay2021}, balanced accuracy, positive predictive value (PPV), F1
\citep{SklearnF1}, matthews correlation coefficient (MCC) \citep{Chicco2020,SklearnMCC} and ROC-AUC \citep{GoogleROC}. Sensitivity, specificity, balanced accuracy and
ROC-AUC are prevalence-independent and at under 4\% positives cannot register
the false-alarm burden, which PPV and MCC can. Reported $p$-values are from
paired tests over the five folds and \textbf{Per-fold results and full confusion matrices are given in the Supplementary Material (Tables~S1--S14).}

\section{Results}\label{sec:results}

\subsection{Zero-shot inference is unusable}

\begin{table*}[htbp]
\centering
\caption{Zero-shot results (Experiment~1) on both renderings. Sens.\ is
sensitivity, Spec.\ specificity, BA balanced accuracy, PPV positive predictive
value, AUC the area under the ROC curve. Almost every configuration collapses:
\nCnM{} predicts every image low-risk, while most others predict nearly every
image high-risk. Balanced accuracy sits at or near 0.50 and PPV never exceeds
0.06, barely above the 3.6--3.8\% base rate. Full confusion matrices are in
Supplementary Tables~S1--S2.}
\label{tab:zs}
\small
\setlength{\tabcolsep}{6pt}
\begin{tabular}{@{}ll rrrrrr r rrrrrr@{}}
\toprule
 & & \multicolumn{6}{c}{Pseudo-coloured} & & \multicolumn{6}{c}{Colourless} \\
\cmidrule(lr){3-8}\cmidrule(lr){10-15}
Set & Strategy & Sens. & Spec. & BA & PPV & MCC & AUC & & Sens. & Spec. & BA & PPV & MCC & AUC \\
\midrule
No Label & \nCM   & 0.82 & 0.21 & 0.52 & 0.04 & 0.01 & 0.51 & & 0.96 & 0.10 & 0.53 & 0.04 & 0.04 & 0.52 \\
No Label & \CnM   & 0.97 & 0.20 & 0.58 & 0.04 & 0.08 & 0.58 & & 1.00 & 0.00 & 0.50 & 0.04 & 0.00 & 0.50 \\
No Label & \CM    & 0.88 & 0.18 & 0.53 & 0.04 & 0.03 & 0.52 & & 0.97 & 0.05 & 0.51 & 0.04 & 0.02 & 0.51 \\
No Label & \nCnM  & 0.00 & 1.00 & 0.50 & 0.00 & 0.00 & 0.50 & & 0.00 & 1.00 & 0.50 & 0.00 & 0.00 & 0.50 \\
\addlinespace
L1 & \nCM   & 0.31 & 0.80 & 0.56 & 0.05 & 0.05 & 0.55 & & 0.92 & 0.13 & 0.53 & 0.04 & 0.03 & 0.52 \\
L1 & \CnM   & 0.97 & 0.20 & 0.58 & 0.04 & 0.08 & 0.58 & & 1.00 & 0.00 & 0.50 & 0.04 & 0.00 & 0.50 \\
L1 & \CM    & 0.54 & 0.62 & 0.58 & 0.05 & 0.06 & 0.57 & & 0.97 & 0.03 & 0.50 & 0.04 & $-0.00$ & 0.49 \\
L1 & \nCnM  & 0.00 & 1.00 & 0.50 & 0.00 & 0.00 & 0.50 & & 0.00 & 1.00 & 0.50 & 0.00 & 0.00 & 0.50 \\
\addlinespace
L2 & \nCM   & 1.00 & 0.00 & 0.50 & 0.04 & 0.00 & 0.50 & & 0.99 & 0.01 & 0.50 & 0.04 & 0.01 & 0.50 \\
L2 & \CnM   & 0.97 & 0.20 & 0.58 & 0.04 & 0.08 & 0.58 & & 1.00 & 0.00 & 0.50 & 0.04 & 0.00 & 0.50 \\
L2 & \CM    & 1.00 & 0.00 & 0.50 & 0.04 & 0.00 & 0.50 & & 0.99 & 0.01 & 0.50 & 0.04 & 0.00 & 0.50 \\
L2 & \nCnM  & 0.00 & 1.00 & 0.50 & 0.00 & 0.00 & 0.50 & & 0.00 & 1.00 & 0.50 & 0.00 & 0.00 & 0.50 \\
\addlinespace
L3 & \nCM   & 0.13 & 0.91 & 0.52 & 0.05 & 0.03 & 0.52 & & 0.46 & 0.56 & 0.51 & 0.04 & 0.01 & 0.51 \\
L3 & \CnM   & 0.97 & 0.20 & 0.58 & 0.04 & 0.08 & 0.58 & & 1.00 & 0.00 & 0.50 & 0.04 & 0.00 & 0.50 \\
L3 & \CM    & 0.26 & 0.85 & 0.55 & 0.06 & 0.05 & 0.55 & & 0.80 & 0.24 & 0.52 & 0.04 & 0.02 & 0.52 \\
L3 & \nCnM  & 0.00 & 1.00 & 0.50 & 0.00 & 0.00 & 0.50 & & 0.00 & 1.00 & 0.50 & 0.00 & 0.00 & 0.50 \\
\bottomrule
\end{tabular}
\end{table*}

\noindent Zero-shot performance (Table~\ref{tab:zs}) is unusable on both renderings and highly sensitive to prompt formulation. The \nCnM{} baseline labels every image
low-risk; under L2 the morphology strategies label almost every image high-risk
(5{,}033 of 5{,}035 on the coloured subset), giving a nominally perfect
sensitivity against zero specificity. Balanced accuracy sits at or near 0.50
almost throughout and MCC near zero, i.e.\ at chance \citep{Chicco2020}. No
prompt design rescues this, which motivates the trained regime.

One row is informative rather than degenerate, and we return to it in
Section~\ref{sec:rq2}: the colour-only strategy \CnM{} uses an \emph{identical}
prompt pair on both renderings, yet reaches balanced accuracy 0.58 and MCC 0.08
on coloured images while collapsing to 0.50 and 0.00 on colourless ones.

\subsection{Training separates the classes; aggregated context does not help}

\begin{table*}[htbp]
\centering
\caption{Trained results (Experiment~2), averaged over five stratified folds.
Sens.\ is sensitivity, Spec.\ specificity, BA balanced accuracy, PPV positive
predictive value, AUC the area under the ROC curve. Best configuration per
rendering in bold. Strategies: \nCM{} morphology only, \CnM{} colour only,
\CM{} both, \nCnM{} no context. Every configuration on the colour-enhanced
rendering exceeds every configuration on the colourless one, on every metric.}
\label{tab:trained}
\small
\setlength{\tabcolsep}{6pt}
\begin{tabular}{@{}ll rrrrrr r rrrrrr@{}}
\toprule
 & & \multicolumn{6}{c}{Pseudo-coloured} & & \multicolumn{6}{c}{Colourless} \\
\cmidrule(lr){3-8}\cmidrule(lr){10-15}
Set & Strategy & Sens. & Spec. & BA & PPV & MCC & AUC & & Sens. & Spec. & BA & PPV & MCC & AUC \\
\midrule
No Label & \nCM   & 0.917 & 0.893 & 0.905 & 0.242 & 0.439 & 0.956 & & 0.823 & 0.821 & 0.822 & 0.155 & 0.307 & 0.888 \\
No Label & \CnM   & 0.917 & 0.891 & 0.904 & 0.239 & 0.437 & 0.954 & & 0.823 & 0.822 & 0.823 & 0.155 & 0.308 & 0.888 \\
No Label & \CM    & 0.917 & 0.893 & 0.905 & 0.242 & 0.439 & 0.956 & & 0.829 & 0.822 & 0.826 & 0.156 & 0.311 & 0.888 \\
No Label & \nCnM  & 0.917 & 0.891 & 0.904 & 0.240 & 0.437 & 0.956 & & 0.829 & 0.824 & 0.826 & 0.157 & 0.312 & 0.890 \\
\addlinespace
L1 & \nCM   & 0.911 & 0.896 & 0.904 & 0.246 & 0.442 & 0.958 & & 0.829 & 0.824 & 0.826 & 0.157 & 0.312 & 0.888 \\
L1 & \CnM   & 0.917 & 0.891 & 0.904 & 0.239 & 0.437 & 0.954 & & 0.823 & 0.822 & 0.823 & 0.155 & 0.308 & 0.888 \\
L1 & \CM    & 0.911 & 0.895 & 0.903 & 0.246 & 0.442 & 0.958 & & 0.824 & 0.824 & 0.824 & 0.156 & 0.309 & 0.886 \\
L1 & \nCnM  & 0.917 & 0.891 & 0.904 & 0.240 & 0.437 & 0.956 & & 0.829 & 0.824 & 0.826 & 0.157 & 0.312 & 0.890 \\
\addlinespace
L2 & \nCM   & 0.906 & 0.887 & 0.896 & 0.231 & 0.424 & 0.952 & & 0.818 & 0.815 & 0.817 & 0.150 & 0.299 & 0.882 \\
L2 & \CnM   & 0.917 & 0.891 & 0.904 & 0.239 & 0.437 & 0.954 & & 0.823 & 0.822 & 0.823 & 0.155 & 0.308 & 0.888 \\
L2 & \CM    & 0.906 & 0.885 & 0.896 & 0.228 & 0.421 & 0.952 & & 0.823 & 0.815 & 0.819 & 0.150 & 0.301 & 0.882 \\
L2 & \nCnM  & 0.917 & 0.892 & 0.904 & 0.240 & 0.438 & 0.956 & & 0.829 & 0.824 & 0.826 & 0.157 & 0.312 & 0.890 \\
\addlinespace
L3 & \nCM   & \textbf{0.917} & \textbf{0.896} & \textbf{0.906} & \textbf{0.248} & \textbf{0.446} & \textbf{0.958} & & 0.835 & 0.831 & 0.833 & 0.164 & 0.322 & 0.894 \\
L3 & \CnM   & 0.917 & 0.891 & 0.904 & 0.239 & 0.437 & 0.954 & & 0.823 & 0.822 & 0.823 & 0.155 & 0.308 & 0.888 \\
L3 & \CM    & 0.911 & 0.895 & 0.903 & 0.244 & 0.440 & 0.956 & & \textbf{0.840} & \textbf{0.834} & \textbf{0.837} & \textbf{0.167} & \textbf{0.328} & \textbf{0.892} \\
L3 & \nCnM  & 0.917 & 0.892 & 0.904 & 0.240 & 0.438 & 0.956 & & 0.829 & 0.824 & 0.826 & 0.157 & 0.312 & 0.890 \\
\bottomrule
\end{tabular}
\end{table*}

\noindent Training changes the picture categorically (Table~\ref{tab:trained}). Averaged over folds the pseudo-coloured model reaches sensitivity 0.914, specificity
0.892, balanced accuracy 0.903, PPV 0.240, MCC 0.437 and ROC-AUC 0.955; the
colourless model reaches 0.827, 0.823, 0.825, 0.157, 0.311 and 0.888. No
configuration collapses. Every metric falls on the colourless subset:
sensitivity by 8.8 points, balanced accuracy by 7.8, MCC by 12.6 and ROC-AUC by
6.7.

Aggregated over all prompt designs, context does nothing. On the pseudo-coloured
subset context-bearing configurations average MCC 0.437 against 0.437 for
no-context ($-0.001$); on the colourless subset, 0.310 against 0.312
($-0.002$). Both differences are far below the fold-to-fold spread of the
baseline itself.

Selected well, context helps, but unevenly. The best pseudo-coloured
configuration, L3 + \nCM{}, reaches MCC $0.446\pm0.016$ against a matched
baseline of $0.438\pm0.014$, but wins in only three folds of five
($+0.008$, $p=0.41$). The best colourless configuration, L3 + \CM{}, reaches
$0.328\pm0.013$ against $0.312\pm0.007$, winning in all five folds ($+0.016$,
$p=0.035$). The asymmetry is the RQ1 result: context is reliable where the
appearance signal is weak and fold-dependent where it is strong.

Label-set means order identically on both renderings and are not monotone in
detail: pseudo-coloured L3 0.440, L1 0.439, No Label 0.438, L2 0.430;
colourless L3 0.318, L1 0.310, No Label 0.309, L2 0.305. L2 is the weakest set
on both renderings and under balanced accuracy as well as MCC, despite sitting
between L1 and L3 in detail. Finally, \CnM{} is identical across label sets
(MCC 0.308 colourless) because it ignores morphology entirely, and it falls
below the no-context baseline (0.312).

\subsection{The colour signal comes from the image, not the prompt}\label{sec:rq2}

\begin{table*}[htbp]
\centering
\caption{Colour in the image crossed with colour in the prompt, trained regime,
averaged over all four label sets (SD over folds where shown). ``Named'' pools
\CnM{} and \CM{}; ``not named'' pools \nCM{} and \nCnM{}. The pattern is
identical under all three metrics.}
\label{tab:rq2}
\small
\begin{tabular}{@{}lccc@{}}
\toprule
 & \multicolumn{2}{c}{Colour in the image} & \\
\cmidrule(lr){2-3}
Colour in the prompt & Present & Absent & Image effect \\
\midrule
\multicolumn{4}{@{}l}{\emph{MCC}} \\
\quad Named         & 0.436 (0.010) & 0.310 (0.013) & $+0.126$ \\
\quad Not named     & 0.438 (0.009) & 0.311 (0.009) & $+0.127$ \\
\quad Prompt effect & $-0.001$ ($p=0.58$) & $-0.001$ ($p=0.75$) & interaction $-0.001$ \\
\addlinespace
\multicolumn{4}{@{}l}{\emph{Balanced accuracy}} \\
\quad Named         & 0.903 (0.019) & 0.825 (0.006) & $+0.078$ \\
\quad Not named     & 0.903 (0.018) & 0.825 (0.007) & $+0.078$ \\
\quad Prompt effect & $-0.001$ ($p=0.83$) & $-0.001$ ($p=0.70$) & interaction $+0.000$ \\
\addlinespace
\multicolumn{4}{@{}l}{\emph{ROC-AUC}} \\
\quad Named         & 0.955 & 0.888 & $+0.067$ \\
\quad Not named     & 0.956 & 0.889 & $+0.067$ \\
\bottomrule
\end{tabular}
\end{table*}

\noindent Table~\ref{tab:rq2} crosses colour in the image with colour in the prompt. The pattern is unambiguous and identical under all three metrics. Moving from
colourless to coloured images gains 0.126 MCC, and gains almost exactly that
whether or not the prompt mentions colour ($+0.126$ named, $+0.127$ not named);
in balanced accuracy both gains are $+0.078$, in ROC-AUC both $+0.067$. Naming
colour changes nothing: $-0.001$ MCC when the colour is there to be named
($p=0.58$) and $-0.001$ when it is not ($p=0.75$), with an interaction of
$-0.001$. The entire colour effect is carried by the image.

The zero-shot regime makes the point more sharply, because there the prompt is
the only thing standing between the encoder and the decision, and the
colour-only prompt pair is identical across renderings. On coloured images it
recovers 175 of 181 positives and, importantly, 972 true negatives (balanced
accuracy 0.58, MCC 0.08, ROC-AUC 0.58). On colourless images the same prompt
labels every one of the 4{,}754 images positive: not a single true negative
(0.50, 0.00, 0.50). The prompt supplies no discriminative content of its own; it
is a query that returns something only when the image can answer it.

This also constrains the colourless result above. Since naming colour has no
main effect there under any metric, the advantage of L3 + \CM{} cannot be a
general benefit of colour context and must be a label-set-specific interaction.

\section{Discussion}\label{sec:discussion}

\subsection{Colour-enhanced images outperform colourless ones on every metric}

\noindent The pseudo-colour overlay, not the prompt, is what separates the two
classes. Once trained, the colour-enhanced rendering beats the colourless one on
every measure we report: balanced accuracy 0.903 against 0.825, sensitivity
0.914 against 0.827, specificity 0.892 against 0.823, PPV 0.240 against 0.157,
ROC-AUC 0.955 against 0.888 and MCC 0.437 against 0.311. The advantage is
therefore not an artefact of one imbalance-sensitive statistic: it appears in
threshold-free ranking (ROC-AUC), at the operating point (balanced accuracy,
MCC), and in both error directions taken separately (sensitivity and
specificity move together). This is what one should expect given
Section~\ref{sec:pseudocolour}: the overlay is a rendering of the labelling rule
itself, already normalised for heart rate, so the encoder is reading the answer
rather than inferring it. The corollary is that the gap should not be read as
the diagnostic value of colour in an ECG; it is the value of showing a model the
output of the algorithm that generated its labels.

\subsection{Text as query, image as evidence}

\noindent  RQ2 admits an unusually clean answer. The same colour prompt is discriminative
when the image contains colour and degenerate when it does not, while the image
contributes its full $\approx0.126$ MCC whether or not the prompt mentions
colour. In a contrastive vision--language model, a prompt naming a visual
attribute appears to function as a \emph{query} over the image rather than as
\emph{evidence} about it: the text selects a direction in the joint embedding
space, and whether anything discriminative lies along that direction is a fact
about the image.

The methodological consequence generalises. A common pattern in context-driven
work is to show that adding an attribute prompt improves performance and
conclude the model is using the attribute. That does not follow, because the
attribute is normally present in image and prompt at once. Where a paired
rendering exists --- images that do and do not contain the attribute, prompt held
fixed --- the confound is removable, and we would encourage this as a routine
control.

It also reframes the negative result on the pseudo-coloured subset. That no
prompt beats the baseline there is not evidence that BiomedCLIP fails to
understand the overlay; the image-side evidence in Table~\ref{tab:rq2} shows the
encoder extracts it well. It is evidence that once extracted, there is nothing
left for text to add --- redundancy in Wolf and Bileschi's \citep{Wolf2006} sense,
demonstrated by direct manipulation rather than inferred from a ceiling.

\subsection{A morphology-independent label bounds what morphology prompts can do}

\noindent  The usefulness of morphological context tracks whether the label-defining cue is visible. Where the overlay presents the rate-normalised nomogram decision, no
strategy improves reliably on the baseline. Where it is removed, morphology is
the only proxy available and the best configuration improves in all five folds.
This is more specific than ``context helps when appearance is weak'': removing
the overlay removes the \emph{answer}, so prompting can contribute only where
the answer must be inferred.

The size of that contribution is bounded by the proxy relation. Because the
source algorithm is deliberately robust to T-wave morphology
\citep{Alahmadi2021}, and because QT-prolonging drugs distort morphology in
drug-specific ways \citep{Vicente2015}, morphological description correlates with
the label largely through drug exposure rather than through the quantity the
label measures. A gain of 0.016 MCC is about what such a proxy should deliver.

This also explains L2, which underperforms every other set on both renderings
despite sitting between L1 and L3 in detail. Prompt structure is part of the
story \citep{Lu2024,An2024}, but the task supplies a sharper one: L2 names
morphological categories --- flat, large, inverted, biphasic, notched --- every one
of which can occur on either side of the nomogram line, so as a set it carries
little information about the quantity being predicted while directing attention
toward shape. L3 smuggles extent back in (``broader and larger than initial
signal''), and it is the language of extent, not the taxonomy, that tracks a
prolonged repolarisation. L2 is not badly phrased so much as badly aimed. The
practical implication is sharper than ``prefer sentences to labels'': a prompt
should name the attribute that \emph{defines} the label, and a clinically
standard taxonomy is no substitute when the taxonomy classifies something else.
Notably, no prompt set tested here names the QT interval or the risk threshold
at all.

It is worth stating plainly what context does when the colour cue \emph{is}
present, since this is the condition in which prompting is most often assumed to
help. The answer is that it can only lose. Against the matched no-context
baseline on colour-enhanced images, the best prompt set (L3 + \nCM) gains 0.008
MCC, 0.002 balanced accuracy and 0.002 ROC-AUC, while the worst (L2 + \CM) loses
0.017, 0.008 and 0.004 --- the downside is roughly twice the upside on MCC and
four times on balanced accuracy. Naming colour specifically, when colour is
there to be named, is exactly neutral (\CnM: MCC 0.437, balanced accuracy 0.904,
ROC-AUC 0.954, against a baseline of 0.437, 0.904 and 0.956). Where the overlay
already encodes the decision, then, the expected return on prompt engineering is
zero and the variance is not: the safest prompt is none. Without colour the two
sides are comparable ($+0.016$ and $-0.013$ MCC; $+0.011$ and $-0.009$ balanced
accuracy), and the best configuration improves in all five folds, so it is only
in the impoverished rendering that context is worth designing at all.

\subsection{Operating-point gains, not better separability}

\noindent One divergence deserves comment rather than suppression. The advantage of the best colourless configuration is $+0.016$ in MCC (5/5 folds, $p=0.035$),
$+0.011$ in balanced accuracy (4/5, $p=0.11$) and $+0.002$ in ROC-AUC (2/5,
$p=0.62$). An effect present at the operating point but not in the
threshold-free ranking is not a contradiction; it is a claim about what the
context does. ROC-AUC measures ranking, which morphology context does not
measurably improve; MCC and balanced accuracy measure performance at the
threshold the balanced probe selects, and there it helps. Context is shifting
the decision boundary rather than improving separability. Accordingly we
describe these as operating-point gains throughout, and we suggest that studies
comparing prompt designs on imbalanced data report both metric families: a
threshold-free metric alone would have shown nothing here, and a
threshold-dependent metric alone would have overstated what was found.

\section{Conclusion}\label{sec:conclusion}

\noindent Once trained, the colour-enhanced rendering outperforms the colourless one on
every metric --- balanced accuracy 0.903 against 0.825, sensitivity 0.914 against
0.827, specificity 0.892 against 0.823, ROC-AUC 0.955 against 0.888 and MCC
0.437 against 0.311 --- confirming that the pseudo-colour overlay carries the
discriminative signal. Aggregated over all prompt designs, context has no
appreciable effect on either rendering (MCC 0.437 vs.\ 0.437 colour-enhanced and
0.310 vs.\ 0.312 colourless; balanced accuracy 0.903 vs.\ 0.904 and 0.825 vs.\
0.826). It is nonetheless two-sided. On colour-enhanced images the best prompt
set gains 0.008 MCC, 0.002 balanced accuracy and 0.002 ROC-AUC over the matched
baseline while the worst loses 0.017, 0.008 and 0.004, so where the overlay
already encodes the decision the downside of context exceeds its upside. Without
colour the two sides are comparable ($+0.016$ and $-0.013$ MCC), the best
configuration improves in all five folds, and context becomes worth designing.

\printcredits

\section*{Declaration of competing interest}
The authors declare that they have no known competing financial interests or
personal relationships that could have appeared to influence the work reported
in this paper.

\section*{Data availability}
The colour-enhanced ECG images are produced and provided by the principal investigator (Alaa Alahmadi), and all data, code and analysis results will be made available on request.

\bibliographystyle{elsarticle-num}
\bibliography{refs}

\clearpage
\includepdf[pages=-]{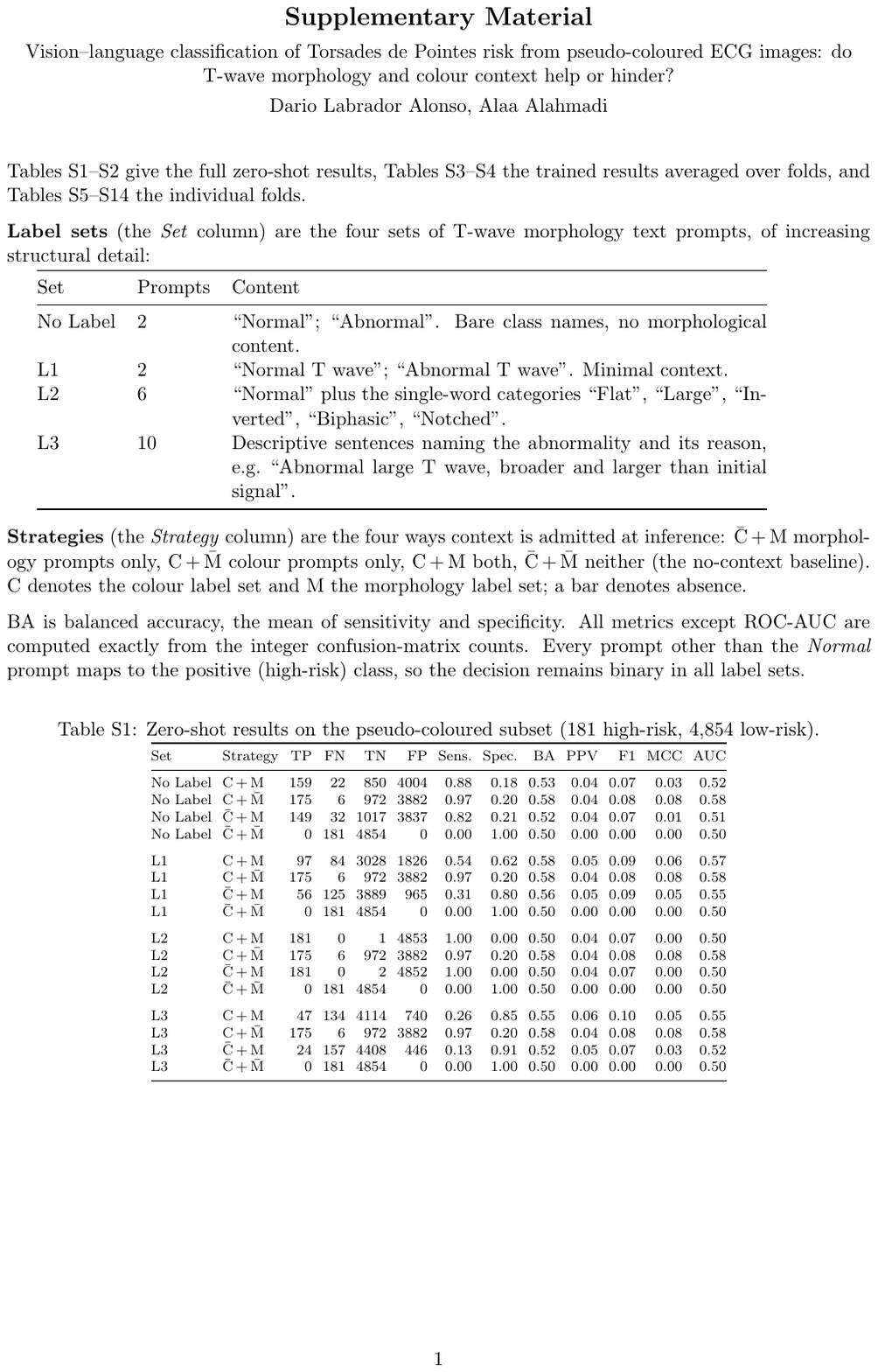}

\end{document}